\documentclass[11pt]{article}
\usepackage{coling2018}
\usepackage{times}
\usepackage{url}
\usepackage{latexsym}
\usepackage{graphicx}
\usepackage{tabularx}
\usepackage{epstopdf}
\usepackage{hyperref}
\usepackage{xstring}

\usepackage{tikzsymbols}
\usepackage{caption,subcaption}
\usepackage{url}

\usepackage{menukeys}

\usepackage{hologo}

\usepackage{fontawesome}

\let\ACMmaketitle=\maketitle
\renewcommand{\maketitle}{\begingroup\let\footnote=\thanks \ACMmaketitle\endgroup}

\title{Towards Enhancing Lexical Resource and Using Sense-annotations of OntoSenseNet for Sentiment Analysis\footnote{ This work was presented at Third Edition of Semantic DeepLearning (SemDeep-3) workshop in The $27^{th}$ International Conference on Computational Linguistics, COLING.}}

\author{Sreekavitha Parupalli, Vijjini Anvesh Rao and Radhika Mamidi  \\
  Language Technologies Research Center, Kohli Center on Intelligent Systems,\\
 International Institute of Information Technology, Hyderabad, India\\
   {\tt \{sreekavitha.parupalli, vijjinianvesh.rao\}@research.iiit.ac.in}\\
   {\tt radhika.mamidi@iiit.ac.in}}

\date{}

\begin{document}
\maketitle
\begin{abstract}
This paper illustrates the interface of the tool we developed for crowd sourcing and we explain the annotation procedure in detail. Our tool is named as `\textit{Parupalli Padajaalam}\footnote{\url{https://github.com/Shreekavithaa/crowd-sourcing}}' which means \textit{web of words by Parupalli}. The aim of this tool is to populate the OntoSenseNet, sentiment polarity annotated Telugu resource. Recent works have shown the importance of word-level annotations on sentiment analysis. With this as basis, we aim to analyze the importance of sense-annotations obtained from OntoSenseNet in performing the task of sentiment analysis. We explain the features extracted from OntoSenseNet (Telugu). Furthermore we compute and explain the adverbial class distribution of verbs in OntoSenseNet. This task is known to aid in disambiguating word-senses which helps in enhancing the performance of word-sense disambiguation (WSD) task(s).  

\end{abstract}

\section{Introduction}

\blfootnote{
\hspace{-0.65cm}  
This work is licensed under a Creative Commons 
Attribution 4.0 International License. 
License details: 
\url{http://creativecommons.org/licenses/by/4.0/}.
}

OntoSenseNet is a lexical resource developed on the basis of Formal Ontology proposed by \cite{otra2015towards}. The formal ontology follows approaches developed by Yaska, Patanjali and Bhartrihari from Indian linguistic traditions for understanding lexical meaning and by extending approaches developed by Leibniz and Brentano in the modern times. Based on this proposed formal ontology, a lexical resource for Telugu language has been developed \cite{parupalli2018enrichment} - OntoSenseNet for Telugu. The resource consists of words tagged with a primary and a secondary sense-types of verbs and sense-classes of adverbs. The sense-identification in OntoSenseNet for Telugu is manually done by experts in the field.

Sentiment analysis deals with the task of determining the polarity of text. To distinguish positive and negative opinions in simple texts such as reviews, blogs, and news articles, sentiment analysis (or opinion mining) is used. There are three ways in which one can perform sentiment analysis : document-level, sentence-level, entity or word-level. These determine the polarity value considering the whole document, sentence-wise polarity, word-wise in some given text respectively \cite{naidu2017sentiment}.

\section{Related Work}

Extensive work has been done in the domain of sentiment analysis for English. We discuss few novel and relevant approaches here.
\cite{esuli2005determining} determines a new method for identifying the opinionated words (subjective terms) in the text based on the quantitative analysis of the glosses of such terms. \cite{gamon2005pulse} present a prototype system for mining topics and sentiment orientation jointly from free text customer feedback. \cite{hatzivassiloglou2000effects} studies the role of adjectives in understanding the subjectivity. \cite{kaji2007building} aims at building a polarity lexicon from massive HTML documents. They propose a model to build a word-level polarity lexicon from the sentence-level polarity annotations.  

\cite{GANGULA18.146} created corpus "Sentiraama" for different domains like movie reviews, song lyrics, product reviews and book reviews in Telugu. Furthermore, his work aims to determine the performance of multi-domain sentiment analysis using reviews from several domains in Sentiraama corpus. \cite{naidu2017sentiment} utilizes Telugu SentiWordNet on the news corpus to perform the task of Sentiment Analysis. \cite{mukku2017actsa} developed a polarity annotated corpus where positive, negative, neutral polarities are assigned to 5410 sentences in the corpus collected from several sources. 

\cite{abburi2016multimodal} proposes an approach to detect the sentiment of a song based on its multi-modality natures (text and audio). The textual lyric features are extracted from the bag of words. By using these features, Doc2Vec generates a single vector for every song. Support Vector Machine (SVM), Naive Bayes (NB) and a combination of both these classifiers are developed to classify the sentiment using the textual lyric features as a part of this work.

\section{Crowd-sourcing Platform}
Crowd-sourcing is an online, distributed problem-solving and production model that has emerged in recent years. Early user input can substantially improve the interaction design. Collecting input from only a small set of participants is problematic in many design situations. To address the above discussed problems, crowd-sourcing is widely adopted by research groups.

\subsection{Add a Word}
Any user can add a word to the resource. The user is prompted to enter the word, it's gloss and a sample sentence which shows the usage of the word. List of words received through this page are manually reviewed before adding to our resource. 

\subsection{User Profiling}
\begin{figure}[!h]
\begin{center}
\includegraphics[width=\columnwidth]{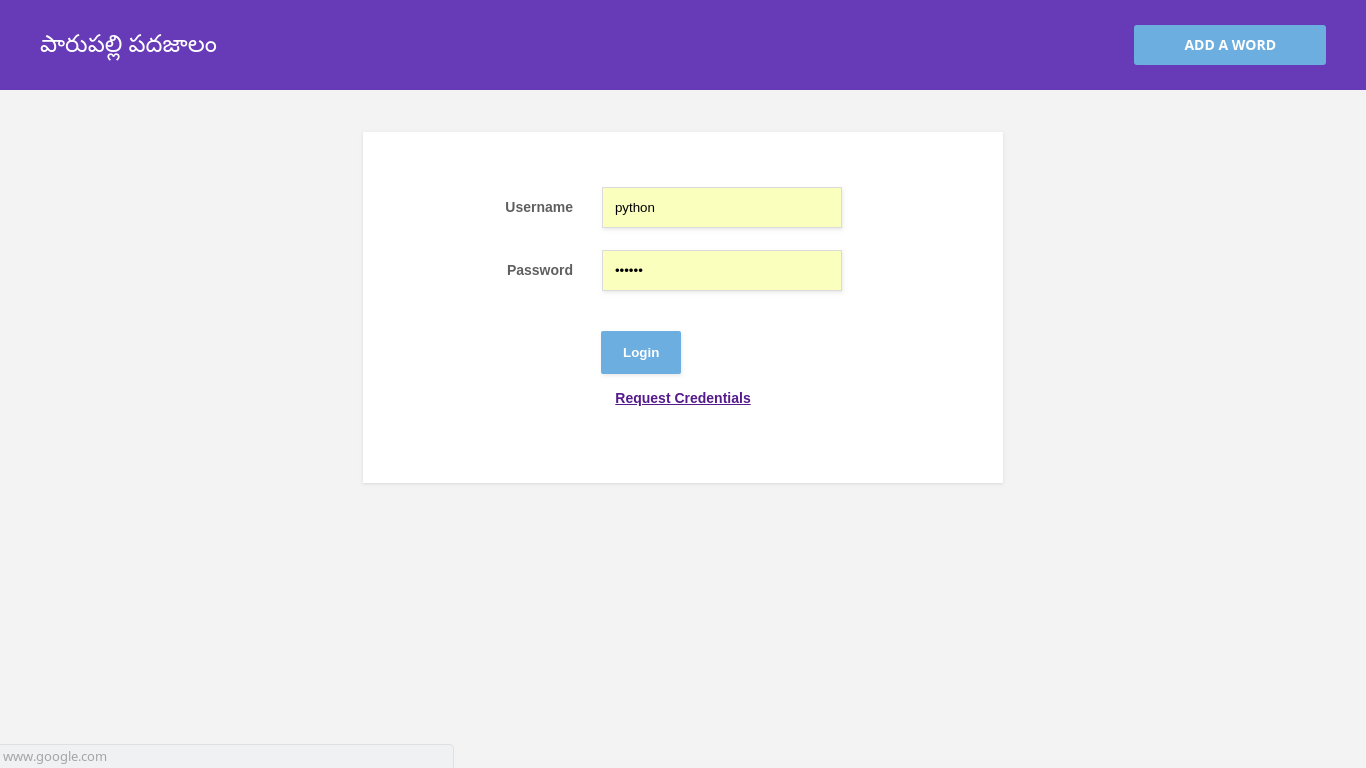} 
\caption{Login page}
\label{fig:login}
\end{center}
\end{figure}

As shown in Figure \ref{fig:login}, any new user who wants to do the annotations must request the login credentials. This is necessary to control the access and avoid unauthentic annotations of the resource. Users are requested to submit their information such as name, email ID, profession, educational background and a score is assigned to the user based on his/her proficiency in Telugu. Score is assigned based on their responses to few questions asked in the request credentials form. This score is used in resolving the conflicts that may arise during annotation. For example, if any word has different tags given by different annotators, the tag given by the annotator with higher score is considered to be accurate. Once the profile is verified, the login credentials are sent to the user through an email.

\subsection{Annotations}
We use this tool to perform two kinds of annotations that are discussed as the part of this thesis. 

\subsubsection{Ontological Sense Annotations}
\begin{figure}[h]
\begin{center}
\includegraphics[width=\columnwidth]{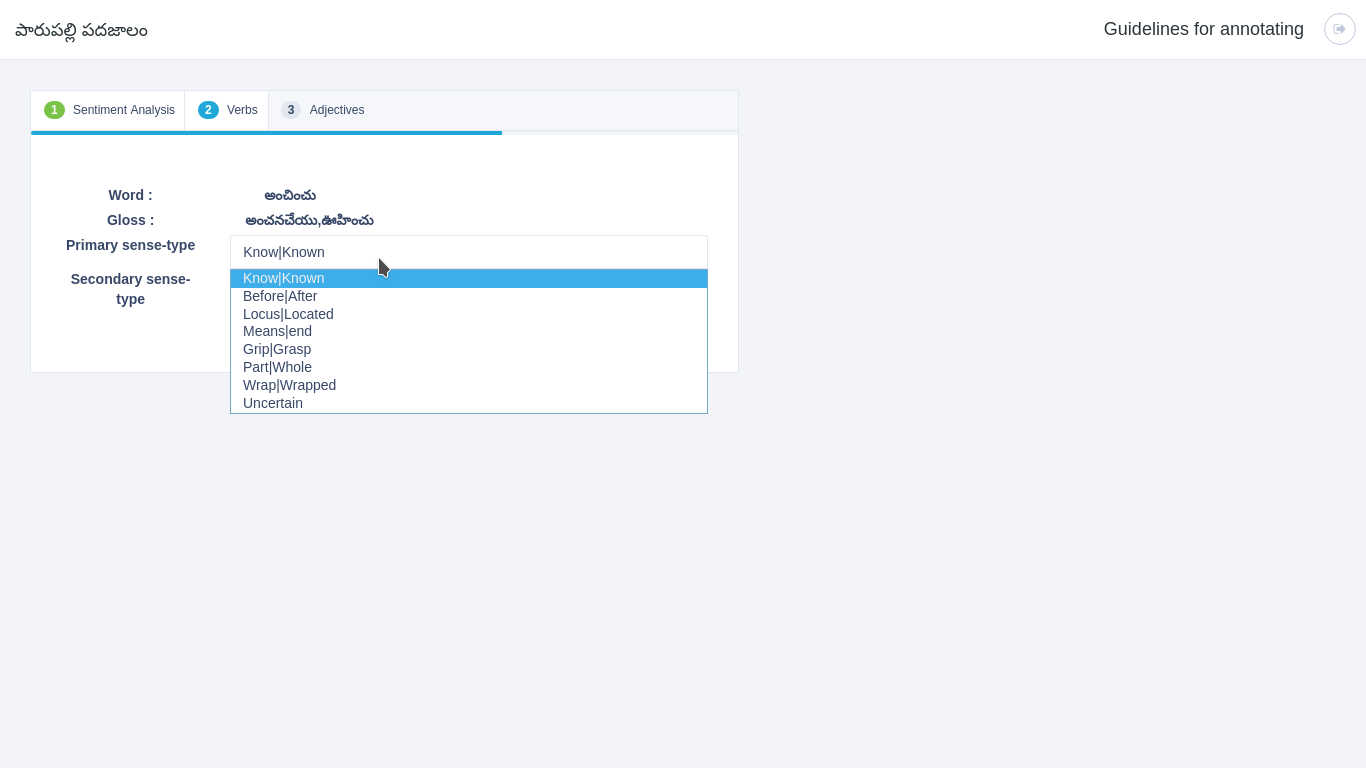} 
\caption{Verb annotation interface with options given to the user}
\label{fig:verb1}
\end{center}
\end{figure}

After logging in, all the users are requested to go through the annotation guidelines before performing the task. These annotation guidelines clearly explain the sense-types and sense-classes proposed. The user is shown a word, its meanings and is prompted to choose the appropriate primary and secondary sense-type of the verbs through the list of options available in the drop down menu as shown in figure \ref{fig:verb1}. Along with the 7 sense-type tags, the user has the liberty  to tag a word(verb) as `uncertain' in case of an unclear judgment. The list of uncertain words are added to the list of the word to-be annotated. Words which are tagged to be uncertain consistently are reviewed and removed from the resource.  
Similar scheme is followed for the adjectives. In case of adjectives, the user could choose to tag the word as any of the 6 defined sense-types or tag the word as `uncertain'.

\subsubsection{Sentiment Polarity Annotations}
In case of polarity annotations, sentiment polarities are classified into 4 labels : positive, negative, neutral and uncertain. Positive and negative labels are given in case of positive and negative sentiments in the word respectively. Uncertain/ambiguous label is given to words which acquire sentiment based on the words it is used along with or it's position in a sentence. Neutral label is given when the word has no sentiment in it. 

\section{Adverbial Class Distribution of Verbs}
\label{v_av}

We have extracted all the Verb-Adverb and Adverb-Verb pairs from the Telugu Wikipedia. In order to acquire these patterns we performed the task of POS tagging\footnote{\url{https://bitbucket.org/sivareddyg/telugu-part-of-speech-tagger}} on Wikipedia corpus. From the extracted pairs, we noticed that there are comparatively more Adverb-Verb pairs than Verb-Adverb pairs which align with the structure of Telugu language\cite{rj2008assessment}. 400 verbs and 445 adverbs are annotated according to the formal ontology that is discussed in \cite{parupalli2018enrichment}. These words formed about 2000 Verb-Adverb and Adverb-Verb pairs. Our aim is to study the adverbial class distribution of verbs in Telugu. \cite{otra2015towards} proves that such annotations help in disambiguating the word senses thus result in improved word-sense disambiguation (WSD) task(s). This is one of the major applications of OntoSenseNet. 

In table \ref{tab:verbadverb}, we show the adverbial class distribution of verbs in Verb-Adverb and Adverb-Verb pairs. Adverbial sense-classes are labeled as columns and sense-types of verbs are labeled as rows. Any cell in the table represents the percentage of a `sense-class' of adverbs that modify a particular `sense-type' of verbs. For example : Column-1 of \ref{tab:verbadverb} means 20.0\% of `spatial'; 13.6\% of `temporal'; 18.8\% of `force' and 24.4\% of `measure' sense-classed of adverbs modify `to know' sense-type of verbs. This shows that majority of the `to know' verbs are primarily modified by adverbs with `measure' sense-class. For example : \textit{cālā anipiṃciṃdi} (feel immensely). `To move', `to do' verbs are primarily modified by `spatial', `force' sense-class of adverbs respectively. Examples are \textit{nerugā māṭlāḍutāḍu}(talk in a straight forward manner), \textit{emoṣanalgā ālocistāḍu} (think emotionally). `Temporal' sense-class of adverbs can modify all the sense-types of verbs. `To be' sense-type of verbs is also significantly modified by `force' sense-class of adverbs. However, `temporal' and `measure' sense-classes also seem to show comparable performance in modifying `to be' sense-type. We can find many such examples in Telugu language.  


\begin{table}[!h]
\begin{center}
\begin{tabular}{| c | c |  c |  c | c | c | c | c |}
\hline
\textbf{}&\textbf{To Know}&\textbf{To Move}&\textbf{To Do}&\textbf{To Have}&\textbf{To Be}&\textbf{To Cut}&\textbf{To Bound}\\
\hline
Spatial & 20.0 \%	& 28.5\%	& 20 \% &	9.5 \%	& 9.5 \% &	8.5\%	& 4.0 \%	\\
\hline
Temporal & 13.6 \% &	22.0 \%	 & 14.6\% &	20.5\%	& 20.5 \% & 4.4\% &	4.4 \% \\
\hline
Force  & 18.8 \% & 21.5 \% & 22.2\%	& 7.2\% & 22.9\% &	6.0\% &	4.1\% \\
\hline
Measure & 24.4\% & 16.5 \% &	19.5\% &	5.2\% &	 20.3 \%	& 7.5\% &3.8\% \\
\hline
\end{tabular}
\end{center}
\caption{Adverb Sense-Class Distribution in Verb-Adverb pairs}
\label{tab:verbadverb}
\end{table}

\section{Validating Importance of Sense-annotations from OntoSenseNet on Sentiment Analysis.}

In an attempt to understand how the sense-type classification of verbs and sense-classes for adverbs could affect sentiment analysis, we perform experiments using sense-annotations from OntoSenseNet(Telugu) as additional features to an existing system as discussed in \cite{2018arXiv180701679P}. We develop a benchmark Word2Vec approach which utilizes averaged word vectors generated from all the words in a review, adopted from Sentiraama corpus\cite{GANGULA18.146}. The constructed review vector is used to determine the benchmark accuracy for sentiment classification. To validate the importance of word-level annotations, the following features are added to our review vectors:

\subsection{Word-level Polarity Features}
SentiWordNet\cite{das2012sentimantics,das2010sentiwordnet,amitava2011dr} is a lexical resource with sentiment polarity annotations. It has 4076 negative and 2135 positive unigrams. \cite{2018arXiv180701679P} shows the importance of word-level annotations on sentiment analysis task. The features they consider are number of positive unigrams from SentiWordNet + their annotated data, number of negative unigrams from SentiWordNet + their annotated data, number of positive bigrams, number of negative bigrams from Sentiraama \cite{GANGULA18.146}. 

\subsection{Additional Features from OntoSenseNet}
Utilizing the sense-annotations from OntoSenseNet resource, we added the following features to our review vectors:

\begin{itemize}

\item Verbs from OntoSenseNet are annotated with 7 sense-type tags namely- To Know, To Move, To Do, To Have, To Be, To Cut, To Bound. We add the frequency of these sense-types in the review, to the averaged word vector of the review. This results in addition of 7 features to the review (feature) vector. 

\item Adverbs from OntoSenseNet are annotated with 4 sense-class tags namely- Spatial, Temporal, Force, Measure adverbs in the review. We add the frequency of these sense-class tags to the review vector. Along with features from obtained from verbs, we add these 4 features. On the whole, we get 11 additional features from OntoSenseNet resource.
\end{itemize}

\subsection{Results}

\ref{tab:acc} shows results of our experiments with various classifiers. K-Nearest neighbor (KNN) classifier shows a huge drop in accuracy after inclusion of the new features. This might be because KNN doesn't differentiate between the features and holds all with equal importance for classification. Hence, it fails to ignore the probable
noisy features among the newly added ones. On the other hand, Random Forest (RF) classifier keeps learning from additional features and is good at ignoring the noisy ones. We observe an interesting trend in the accuracies i.e. performance of Linear SVM keeps decreasing and at the same time performance of Gaussian SVM keeps increasing. This shows loss of linear separability. On the whole, we find Neural Network (NN) to be best performing classifier when averaged over repeated trials. However, repeated trials of the experiment show high variance. \ref{tab:nn} shows in detail the precision, recall, and f1-scores of Neural Network's performance with two hidden layers of size 100 and 25 and input vectors of 200 dimensions without additional features. The increment in accuracy over addition of features extracted from OntoSenseNet validate our hypothesis that OntoSenseNet \textit{does} contain semantic knowledge valuable to the task of sentiment analysis. 
\begin{table}[!h]
\begin{center}
\begin{tabular}{| c | c |  c |  c | c | }
\hline
\textbf{}&\textbf{Word2Vec}&\textbf{+word-level}&\textbf{+ OntoSenseNet }&\textbf{+ Both}\\ &&\textbf{polarity features}&\textbf{features}&\\
\hline
Linear SVM & 81.59 \%	& 70.64\%	& 78.10 \% &	76.11 \%\\
\hline
Gaussian SVM & 48.25 \% &	67.66 \%	 & 66.16\% &	73.63\%	\\
\hline
Random Forest & 74.62 \% & 75.12 \% & 77.61\%	& 75.62\% \\
\hline
Neural Network & 81.09\% & 75.62 \% &	\textbf{83.08}\% &	81.09\% \\
\hline
K-Nearest Neighbor & 81.09\% & 62.68 \% &	65.17\% &	68.15\% \\
\hline
\end{tabular}
\end{center}
\caption{Accuracy for various classifier with different features.}
\label{tab:acc}
\end{table}

\begin{table}[!h]
\begin{center}
\begin{tabular}{| c | c |  c |  c | c | }
\hline
\textbf{}&\textbf{Word2Vec}&\textbf{+word-level}&\textbf{+ OntoSenseNet }&\textbf{+ Both}\\ &&\textbf{polarity features}&\textbf{features}&\\
\hline
Precision & 0.820 & 0.760	& 0.833 &	0.811\\
\hline
Recall & 0.813 & 0.753 & 0.829 &0.811\\
\hline
F-Measure & 0.810& 0.753 & 0.829& 0.810 \\
\hline

\end{tabular}
\end{center}
\caption{Precision, Recall and F1-scores for Neural Network with different features.}
\label{tab:nn}
\end{table}

\section{Conclusion}
This paper presents the tool developed for crowd-sourcing the annotations that are yet to-be done. In the development of OntoSenseNet only 1673 adjectives out of 11,000 were annotated. Rest of these adjectives will be annotated through crowd-sourcing approach. Additional verbs extracted from WordNet also need annotations to be done. In sentiment analysis task, most of the unigram annotations are done through crowd-sourcing approach. Before this tool is developed, annotations are crowd-sourced using Amazon Mechanical Turk (MTurk) \footnote{\url{https://www.mturk.com}}. The bigrams (verb, adverb pairs) discussed in section \ref{v_av} are yet to be annotated through the tool. The adverbial class distribution of verbs is extracted from Telugu Wikipedia that is aimed to improvise WSD tasks. We present the insights obtained from the statistics. We validate our hypothesis that features extracted from OntoSenseNet carry relevant information that is useful for sentiment analysis through machine learning approaches. 

\section{Acknowledgments}
I want to thank Abhilash Reddy for his help in developing the crowd sourcing tool. I want to thank Mrs. Vijaya Lakshmi Kiran Kumar for her continuous support and encouragement.  

\bibliographystyle{acl}
\bibliography{coling2018}
\end{document}